\documentclass[conference]{IEEEtran}
\IEEEoverridecommandlockouts
\usepackage{amsmath,amsfonts}
\usepackage{array}
\usepackage{textcomp}
\usepackage{stfloats}
\usepackage{url}
\usepackage{amsmath, amssymb, amsthm}
\usepackage{verbatim}
\usepackage{graphicx}
\usepackage{amssymb}
\usepackage{hyperref}
\usepackage{graphicx}
\usepackage{amsmath}
\usepackage{longtable}
\usepackage{algorithm} 
\usepackage{algpseudocode} 
\usepackage{mathrsfs}
\usepackage{subcaption}
\usepackage{mathtools}
\usepackage{pifont}
\usepackage{color}
\usepackage{lineno}
\usepackage{graphicx}  
\usepackage{makecell} 
\usepackage{pdflscape}
\usepackage{adjustbox}
\usepackage[utf8]{inputenc}
\usepackage{tabularx}
\usepackage{blindtext}
\usepackage{longtable}
\usepackage{lscape}
\usepackage{amsthm}
\usepackage{graphicx}
\usepackage{subcaption} 

\usepackage{setspace}
\usepackage{notoccite} 
\usepackage{lscape} 
\usepackage{mwe}
\usepackage{booktabs}
\usepackage{amsthm}

\theoremstyle{definition}

\theoremstyle{definition}

\newcommand{\RNum}[1]{\lowercase\expandafter{\romannumeral #1\relax}}
\newcommand{\RNumU}[1]{\uppercase\expandafter{\romannumeral #1\relax}}
\usepackage[numbers]{natbib}

\def\BibTeX{{\rm B\kern-.05em{\sc i\kern-.025em b}\kern-.08em
    T\kern-.1667em\lower.7ex\hbox{E}\kern-.125emX}}
\begin{document}

\title{\title{Robust Dual-Model Collaborative Random Vector Functional Link Network}
}

\author{
\IEEEauthorblockN{A. Quadir}
\IEEEauthorblockA{
\textit{Department of Mathematics} \\
\textit{Indian Institute of Technology Indore}\\
mscphd2207141002@iiti.ac.in}
\and
\IEEEauthorblockN{A. Rahaman}
\IEEEauthorblockA{
\textit{Department of Mathematics} \\
\textit{Indian Institute of Technology Indore}\\
phd2401141001@iiti.ac.in}
\and
\IEEEauthorblockN{Mushir Akhtar}
\IEEEauthorblockA{
\textit{Department of Mathematics} \\
\textit{Indian Institute of Technology Indore}\\
phd2101241004@iiti.ac.in}
\and
\IEEEauthorblockN{M. Tanveer\textsuperscript{*}\thanks{\textsuperscript{*}Corresponding author}}
\IEEEauthorblockA{
\textit{Department of Mathematics} \\
\textit{Indian Institute of Technology Indore}\\
mtanveer@iiti.ac.in}
}

\maketitle

\begin{abstract}
Random vector functional link (RVFL) networks are lightweight and fast neural models that offer efficient training and strong generalization through randomized hidden-layer weights and direct input-output connections. However, conventional RVFL models are sensitive to noisy labels, outliers, and imbalanced data, which limits their performance in real-world applications. To address these challenges, we propose the kernel risk-sensitive mean p-power based RVFL (KRPRVFL) model, which integrates the computational efficiency of RVFL with the robustness of the kernel risk-sensitive mean p-power (KRP) criterion. By replacing the standard least-squares objective with a KRP-based loss, KRPRVFL adaptively reduces the influence of corrupted or unreliable samples during training, resulting in improved stability and generalization. Additionally, a collaborative learning mechanism is introduced to enable adaptive interaction among model components, further enhancing robustness in complex and noisy environments. The proposed framework also leverages kernel-induced feature mapping to capture nonlinear relationships without requiring explicit hidden-layer selection, maintaining both efficiency and scalability. Extensive experiments on UCI and KEEL benchmark datasets demonstrate that KRPRVFL consistently outperforms baseline models in terms of accuracy, robustness, and statistical significance, highlighting its effectiveness as a fast, scalable, and reliable solution for challenging classification tasks. The code and supplementary material of the paper can be accessed using the following link: \url{https://github.com/mtanveer1/KRPRVFL}.
\end{abstract}
\section{Introduction}
\IEEEPARstart{I}{n} recent years, deep learning has emerged as a dominant paradigm in artificial intelligence and has achieved remarkable success across a wide range of applications \cite{10716703, quadir2026hypergraph}. Despite its strong representation and learning ability, deep learning models typically rely on complex architectures and involve a large number of hyperparameters, which makes the training process computationally expensive and time-consuming \cite{quadir2024granular}. To address these limitations, \citet{pao1994learning} introduced the random vector functional link (RVFL) network, which offers a lightweight alternative with a simple architecture and a reduced set of tunable parameters, enabling rapid learning without the need for iterative parameter updates \cite{sajid2025gb}. RVFL is a shallow feed-forward randomized neural network in which the hidden-layer weights are randomly generated and kept fixed during training. A distinctive characteristic of RVFL is the presence of direct connections between the input and output layers \cite{akhtar2025towards, zhang2016comprehensive}. These shortcut connections provide an implicit form of regularization, thereby enhancing learning stability and improving generalization performance compared to conventional randomized neural networks \cite{quadir2025randomized, sajid2025wave}.  

To further enhance the generalization ability of the standard RVFL framework, a number of improved variants have been proposed, aiming to increase robustness and practical effectiveness \cite{malik2023random, tanveer2025robust}. RVFL treats all training samples uniformly, which can make the model vulnerable to the presence of noise and outliers \cite{quadir2024multiview, quadir2026garfln}. To mitigate this drawback, an intuitionistic fuzzy RVFL (IFRVFL) model is introduced in \cite{malik2022alzheimer}, where fuzzy membership and non-membership functions are employed to assign intuitionistic fuzzy scores to individual samples, thereby reducing the influence of unreliable data. Moreover, in conventional RVFL, the input features are mapped into a randomized feature space, which may introduce instability in the learned representation. To alleviate this issue, \citet{zhang2019unsupervised} integrated a sparse autoencoder with $\ell_{1}$-norm regularization into the RVFL framework, leading to the development of the SP-RVFL model. By enforcing sparsity in the learned representations, this approach reduces the adverse effects introduced by random feature mapping and promotes more stable and informative feature extraction. Recently, complex-valued extensions of RVFL have been introduced to address limitations of real-valued models in complex signal processing \cite{liu2025complex}. In particular, CRVFL and its augmented variants effectively exploit complex-valued statistics and correlations between real and imaginary components, achieving improved performance and computational efficiency in complex-valued learning tasks while preserving the fast training advantage of RVFL \cite{sajid2025rvfl}. Although these extended RVFL variants enhance robustness to a certain extent, they remain inadequate for effectively addressing classification tasks involving heavily contaminated or noisy data commonly encountered in real-world applications \cite{quadir2024intuitionistic}.

The kernel risk-sensitive mean (p)-power (KRP) criterion is effective in reproducing kernel Hilbert spaces (RKHS) for robust learning \cite{perla2025short}. By combining kernel-based nonlinear mapping with a risk-sensitive formulation, it improves resilience to noise, outliers, and data imbalance \cite{quadir2025twin, quadir2025trkm, quadir2024one}. Initially developed for recursive kernel adaptive filtering and later extended to complex-valued learning \cite{huang2021complex}, it has broad applicability. Additionally, a kernel-based RVFL model (KERVFL) avoids explicit hidden-layer size selection \cite{chakravorti2020non}. However, kernel methods suffer from high computational cost and scalability issues, and integrating KRP into RVFL remains underexplored.

Motivated by the limitations of conventional RVFL networks in handling noisy, imbalanced, or corrupted data, we propose the kernel risk-sensitive mean p-power based RVFL (KRPRVFL) model, which integrates the efficiency of RVFL with the robustness of the KRP criterion. Unlike standard RVFL, which relies on a least-squares objective and treats all samples equally, KRPRVFL employs a risk-sensitive KRP-based loss function to adaptively suppress the influence of outliers and mislabeled samples during training. This ensures that the model focuses on the underlying data distribution, resulting in improved generalization and stability in noisy environments. Furthermore, KRPRVFL incorporates a collaborative learning mechanism, enabling dynamic interaction among the model components to refine predictions and adapt to complex data patterns. By leveraging kernel-induced feature mapping, KRPRVFL captures nonlinear relationships in the input space without requiring manual selection of hidden-layer size or complex iterative training procedures. The proposed KRPRVFL model provides a fast, scalable, and robust framework for classification tasks, effectively combining the computational simplicity of RVFL with the resilience and adaptability of kernel-based risk-sensitive learning.

The key highlights of this paper can be encapsulated as follows:
\begin{enumerate}
    \item The proposed KRPRVFL model combines the fast and lightweight architecture of RVFL networks with the robust kernel risk-sensitive mean p-power (KRP) criterion, enabling effective handling of noisy, corrupted, and imbalanced data.
    \item By replacing the conventional least-squares objective with a KRP-based loss, the model adaptively suppresses the influence of outliers and mislabeled samples, improving generalization and stability under challenging learning conditions.
    \item KRPRVFL uses collaborative learning and kernel mapping to capture nonlinear relationships, improving accuracy and robustness without needing to set hidden-layer size.
    \item Extensive experiments on UCI and KEEL benchmark datasets demonstrate that the proposed KRPRVFL consistently outperforms baseline models in terms of accuracy and statistical significance.
\end{enumerate}

\section{Related work}
In this section, we first establish the notations used throughout the paper and then provide an overview of the RVFL model.

\subsection{Notations}
Let the training dataset be represented as $\mathcal{X} = \{(x_i, y_i) \mid i = 1, 2, \dots, n\}$, where $x_i \in \mathbb{R}^{1 \times m}$ denotes the input feature vector and $y_i \in \{+1, -1\}$ is the corresponding target label. $n$ is the total number of training samples and $m$ is the number of features. The transpose operator is denoted by $(\cdot)^T$. The matrices of all input and output samples are defined as $X = [x_1^T, x_2^T, \dots, x_n^T]^T$ and $Y = [y_1^T, y_2^T, \dots, y_n^T]^T$, respectively.

\subsection{Random Vector Functional Link (RVFL) Network}
The RVFL network is a single-layer feedforward network with random, fixed input-to-hidden weights and shortcut connections from input to output, enhancing generalization. The hidden-layer output matrix $H_1 \in \mathbb{R}^{n \times N}$, with $N$ hidden nodes, is computed as:
\begin{align}
H_1 = \phi(X W_1 + b_1),
\end{align}
where $W_1 \in \mathbb{R}^{m \times N}$ is a randomly initialized weight matrix, $b_1 \in \mathbb{R}^{n \times N}$ is the bias matrix, and $\phi$ is the activation function.

The combined feature matrix $H_2$, which concatenates the input and hidden-layer outputs, is defined as $H_2 = [X ; H_1]$. The predicted output $\hat{Y}$ is then given by:
\begin{align}
H_2 \beta = \hat{Y},
\end{align}
where $\beta \in \mathbb{R}^{(m+N) \times 1}$ denotes the output weight matrix. Training the RVFL involves solving the following regularized least-squares optimization problem:
\begin{align}
\beta_{\min} = \arg\min_\beta \frac{\mathcal{C}}{2} \|H_2 \beta - Y\|^2 + \frac{1}{2} \|\beta\|^2,
\end{align}
where $\mathcal{C} > 0$ is a regularization parameter.

The closed-form solution for $\beta$ depends on the dimensions of $H_2$ relative to the number of samples $n$:
\begin{align}
\beta_{\min} =
\begin{cases}
H_2^T \left(H_2 H_2^T + \frac{1}{\mathcal{C}} I\right)^{-1} Y, & n < m + N, \\
\left(H_2^T H_2 + \frac{1}{\mathcal{C}} I\right)^{-1} H_2^T Y, & n \ge m + N,
\end{cases}
\end{align}
where $I$ is an identity matrix of appropriate size. 

\section{The Proposed Robust Dual-Model Collaborative Random Vector Functional Link Network}
In the presence of label noise, the output weights learned by conventional RVFL networks become highly sensitive to corrupted labels, leading to degraded robustness and generalization performance. To address this issue, kernel risk-sensitive mean p-power based RVFL (KRPRVFL) is proposed, in which the output weights are learned using the KRP criterion. By replacing the conventional least-squares objective, KRP-RVFL effectively suppresses the influence of mislabeled samples and achieves robust learning under label noise environments. 

The KRP metric, defined in RKHS, emphasizes deviations to reduce the impact of outliers, enhancing RVFL robustness in noisy environments.
In kernel-induced feature spaces, higher-order statistical relationships in the original input space can be equivalently represented using second-order statistics. Let $\alpha$ and $\beta$ denote two random variables. Their dependency can be quantified in RKHS via a kernel-based correlation measure \cite{yi2023identifying, zhang2021resetting}, which is expressed as:
\begin{align}
    \mathcal{V}(\alpha, \beta) = \mathbb{E}\left[\langle \phi(\alpha), \phi(\beta) \rangle_{\mathcal{H}}\right]
= \int \langle \phi(\alpha), \phi(\beta) \rangle_{\mathcal{H}} , dF_{\alpha,\beta}(\alpha,\beta),
\end{align}
where $\mathbb{E}[\cdot]$ denotes the expectation operator and $F_{\alpha,\beta}(\alpha,\beta)$ represents the joint probability distribution of $\alpha$ and $\beta$. The mapping $\phi(\cdot)$ is implicitly defined through a Mercer kernel $\kappa_\sigma(\cdot)$, which projects data from the input space into an RKHS $\mathcal{H}$ equipped with the inner product $\langle \cdot, \cdot \rangle_{\mathcal{H}}$. This mapping satisfies:
\begin{align}
    \langle \phi(\alpha), \phi(\beta) \rangle_{\mathcal{H}} = \kappa_\sigma(\alpha, \beta).
\end{align}
Based on the above formulation, the KRP criterion is constructed to quantify the similarity between random variables in RKHS while incorporating risk sensitivity, which forms the foundation for robust output-weight learning in the proposed KRPRVFL framework. The KRP criterion is defined as follows: 
\begin{align}
    \mathcal{L}_{\mu, p, \sigma}(\alpha, \beta) & = \frac{1}{\mu} \mathbb{E} \left[ \exp \left(2^{-\frac{1}{p}} \mu \| \phi(\alpha) - \phi(\beta) \|^p_{\mathcal{H}}  \right) \right] \nonumber \\
    & = \frac{1}{\mu} \mathbb{E} \left[ \exp \left(2^{-\frac{1}{p}} \mu \left( \| \phi(\alpha) - \phi(\beta) \|^p_{\mathcal{H}} \right)^{\frac{p}{2}}  \right) \right] \nonumber \\
    & = \frac{1}{\mu} \int \exp \left( \mu (1 - \kappa_{\sigma}(\alpha, \beta))^{\frac{p}{2}} \right)dF_{\alpha,\beta}(\alpha,\beta),
\end{align}
where $\alpha$ and $\beta$ denote arbitrary random variables. The parameter $\mu > 0$ regulates the degree of risk sensitivity in the loss function, while $p > 0$ determines the order of the deviation penalty. Throughout this work, the kernel mapping is instantiated using a Gaussian Mercer kernel with a bandwidth parameter $\sigma > 0$, defined as:
\begin{align}
    \kappa_{\sigma}(\alpha,\beta) = \exp\left(
-\frac{(\alpha - \beta)^{2}}{2\sigma^{2}}
\right).
\end{align}
The joint probability distribution of $(\alpha,\beta)$ is generally inaccessible. Instead, one only observes a finite set of $n$ paired samples $(\alpha_i, \beta_i)_{i = 1}^n$. Using the empirical approximation, the KRP criterion is formulated as:
\begin{align}
    \mathcal{L}_{\mu,p,\sigma}(\alpha,\beta) = \frac{1}{n\mu} \sum_{i=1}^{n} \exp \left(\mu \left(1-\kappa_{\sigma}(\alpha_i,\beta_i)\right)^{\frac{p}{2}} \right).
\end{align}

From this perspective, the KRP criterion can be interpreted as a similarity evaluation mechanism between two sample sequences $[\alpha_1, \alpha_2, \ldots, \alpha_n]$ and $[\beta_1, \beta_2, \ldots, \beta_n]$ in the kernel-induced feature space. The learning task aims to minimize the empirical KRP loss between the true label matrix $Y$ and the RVFL output, which can be expressed as:
\begin{align}
    \mathcal{L}_{\mu, p, \sigma} (Y, HW) = \underset{W}{\arg \min} \sum_{i=1}^n \exp (\mu (1- \kappa_\sigma (y_i, \hat{y}_i))^{\frac{p}{2}}),
\end{align}
where $y_i$ and $\hat{y}_i$ denote the true label and the predicted output of the $i^{th}$ sample $x_i$, respectively. The predicted output $\hat{y}_i$ is obtained from the RVFL model as:
\begin{align}
    \hat{y}_i = h_i W,
\end{align}
where $h_i \in \mathbb{R}^{N}$ represents the $i^{th}$ row of the RVFL hidden-layer feature matrix and $W$ denotes the output weight matrix.

To further control model complexity and prevent overfitting, a regularization term is incorporated into the objective function. Consequently, the final optimization problem of the proposed KRPRVFL model is formulated as:
\begin{align}
    \underset{W}{\arg \min} \left( \mathcal{D} \|W\|^2  + \frac{1}{n\mu} \sum_{i=1}^n \left(\mu (1 - \kappa_\sigma(y_i, h_iW))^{\frac{p}{2}} \right) \right).
\end{align}

To simplify the subsequent derivations, the overall objective function is denoted by $\psi(W)$, defined as
\begin{align}
    \psi(W) = \mathcal{D} \|W\|^2 + \frac{1}{n\mu} \sum_{i=1}^{n} \exp \left( \mu \left(1-\kappa_{\sigma}(y_i, h_iW)\right)^{[p/2]} \right).
\end{align}
For notational convenience, we further introduce the auxiliary variable $\upsilon _i = 1 - \kappa_{\sigma}(y_i, h_iW),$ which characterizes the kernel-based deviation between the true label and the predicted output of the $i^{th}$ sample. Based on this definition, the gradient of $\psi(W)$ with respect to the output weight matrix $W$ can be derived as follows:
\begin{align}
\label{EQQ:10}
    \frac{\partial \psi(W)}{\partial W} = 2 \mathcal{D}W - \frac{p}{2n\sigma^2}H^T \Omega (Y - HW),
\end{align}
where $\Omega \in \mathbb{R}^{n \times n}$ denotes a diagonal weighting matrix defined in Eqs. \eqref{EQQ:11} and \eqref{EQQ:12}. The elements on its diagonal are determined by the contribution of each training sample to the overall loss function, thereby reflecting the relative influence of individual samples during the optimization process. Specifically, the $i^{th}$ diagonal entry is computed as:
\begin{align}
\label{EQQ:11}
    \Omega_{ii} = \kappa_{\sigma}(y_i, h_i W) \left(\upsilon_i\right)^{\frac{p-2}{2}} \exp \left( \mu \left(\upsilon_i\right)^{\frac{p}{2}}
\right),
\end{align}
and 
\begin{align}
\label{EQQ:12}
\Omega =
\begin{bmatrix}
    \kappa_{\sigma}(y_1, h_1 W) \cdot (\upsilon_1)^{\frac{p-2}{2}} \\
    \exp \left(\mu (\upsilon_1)^{\frac{p}{2}}\right) & & \\    
     & \ddots &  \\
    & & \kappa_{\sigma}(y_N, h_N W)\cdot(\upsilon_n)^{\frac{p-2}{2}} \\
    &  & \cdot \exp \left(\mu (\upsilon_n)^{\frac{p}{2}}\right)
\end{bmatrix}.
\end{align}
By equating the gradient in \eqref{EQQ:10} to zero, a closed-form solution for the output weight matrix $W$ can be derived as:
\begin{align}
\label{EEQ:13}
    W = (\mu'I + H^T \Omega H)^{-1}H^T \Omega Y,
\end{align}
where $\mu' = \frac{4 \mathcal{D} n\sigma^{2}}{p}$ serves as the regularization-related scalar. To improve efficiency, an equivalent form of \ref{EEQ:13} is derived using matrix inversion identities, given by
\begin{align}
\label{EEQ:14}
    W = H^T (\mu' I + \Omega H H^T)^{-1}\Omega Y.
\end{align}
The preferred formulation depends on the sample size $n$ and feature dimension $N+L$: use \eqref{EEQ:13} if $n > L+N$, and (\eqref{EEQ:14}) if $n < L+N$. The update rule in \eqref{EEQ:14} depends on the current $W$, allowing it to be viewed as a fixed-point mapping:
\begin{align}
    W = \mathcal{F}(W, H, Y) = (\mu' I + H^T \Omega H)^{-1} H^T \Omega Y.
\end{align}
The output weight matrix at the ($t^{th}$ iteration is obtained by applying the mapping $\mathcal{F}(\cdot)$ to the estimate from the previous iteration, which yields:
\begin{align}
    W_t = \mathcal{F}(W_{t-1}, H, Y).
\end{align}

The complete procedural steps for the KRPRVFL model are summarized in Algorithm \ref{alg:KRPRVFL}.
\begin{algorithm}[ht!]
\caption{KRPRVFL Algorithm}
\label{alg:KRPRVFL}
\begin{algorithmic}[1]
\State \textbf{Input:} Training data $X$ with labels $Y$; regularization parameters $\mathcal{D}$; kernel bandwidth $\sigma$; convergence threshold $\tau$; maximum number of iterations $T$; risk-sensitivity parameter $\mu$; power parameter $p$.
\State \textbf{Output:} Output weights $W$.
\State Generate the hidden-layer feature matrix $\mathcal{B} = \gamma(X \mathcal{O} + b)$, where $\mathcal{O}$ is the randomly initialized weights matrix and $b$ is the bias vector, and $\gamma$ is an activation function.
\State Find the enhanced features using $H = [X~\mathcal{B}]$.
\For{t = 1, 2, \ldots, T}
    \State Compute the sample-adaptive weighting matrix using Eqs. \eqref{EQQ:11} and \eqref{EQQ:12}.
    \State Update the output weight matrix $W$ using Eq. \eqref{EEQ:13} or Eq. \eqref{EEQ:14}.
    \If{$\lVert W_t - W_{t-1} \rVert^2 < \tau$}
        \State \textbf{break}
    \EndIf
\EndFor
\end{algorithmic}
\end{algorithm}

\begin{table*}[ht!]
\centering
    \caption{Performance comparison of the proposed KRPRVFL model along with the baseline models for UCI and KEEL datasets.}
    \label{Classification performance UCI and KEEL}
    \resizebox{0.8\linewidth}{!}{
\begin{tabular}{lccccccc}
\hline
 Dataset & RVFL \cite{pao1994learning} & ELM \cite{huang2006extreme} & GB-RVFL \cite{sajid2025gb} & GE-GB-RVFL \cite{sajid2025gb} & CRVFL \cite{liu2025complex} & ACRVFL \cite{liu2025complex} & KRPRVFL$^{\dagger}$ \\
 \hline
bank & 89.17 & 87.31 & 88.43 & 88.95 & 88.28 & 88.8 & 89.31 \\
blood & 76.44 & 72.14 & 76 & 77.33 & 75.56 & 77.33 & 77.33 \\
breast\_cancer & 72.09 & 74.42 & 77.91 & 65.12 & 79.07 & 74.42 & 74.42 \\
breast\_cancer\_wisc\_prog & 75 & 73.33 & 73.33 & 73.33 & 66.67 & 66.67 & 63.33 \\
bupa or liver-disorders.csv & 65.38 & 65.38 & 54.81 & 56.73 & 94.95 & 77.24 & 64.42 \\
checkerboard\_Data.csv & 85.94 & 85.98 & 87.02 & 87.02 & 62.48 & 71.5 & 87.02 \\
chess\_krvkp & 90.41 & 90.2 & 91.45 & 89.16 & 93.95 & 94.06 & 97.81 \\
cleve.csv & 81.11 & 80 & 75.56 & 82.22 & 73.79 & 79.79 & 84.44 \\
conn\_bench\_sonar\_mines\_rocks & 74.6 & 71.43 & 74.6 & 68.25 & 65.08 & 66.67 & 84.13 \\
credit\_approval & 84.62 & 82.3 & 77.88 & 80.77 & 83.65 & 84.13 & 83.17 \\
crossplane150.csv & 81.11 & 81.11 & 86.67 & 73.33 & 84.68 & 73.79 & 95.56 \\
cylinder\_bands & 72.73 & 70.67 & 59.74 & 64.29 & 70.78 & 66.23 & 74.03 \\
ecoli-0-1-4-6\_vs\_5.csv & 98.81 & 98.81 & 98.81 & 98.81 & 87.14 & 86.74 & 95.83 \\
ecoli-0-1-4-7\_vs\_5-6.csv & 90 & 96 & 94 & 94 & 72.9 & 70.66 & 96 \\
fertility & 90 & 80 & 86.67 & 90 & 90 & 90 & 90 \\
haber.csv & 76.09 & 78.26 & 78.26 & 77.17 & 75.46 & 85.85 & 78.26 \\
haberman.csv & 76.09 & 76.09 & 78.26 & 77.17 & 71.58 & 83.33 & 78.26 \\
haberman\_survival & 76.09 & 78.26 & 78.26 & 77.17 & 82.61 & 82.61 & 77.17 \\
heart\_hungarian & 72.78 & 72.65 & 74.16 & 74.16 & 75.28 & 76.4 & 78.65 \\
hepatitis & 72.34 & 70.11 & 72.34 & 72.34 & 76.6 & 82.98 & 82.98 \\
hill\_valley & 68.41 & 67.31 & 59.07 & 60.71 & 53.3 & 54.12 & 72.25 \\
horse\_colic & 83.78 & 80.18 & 72.97 & 76.58 & 76.58 & 77.48 & 85.59 \\
ionosphere & 82.79 & 82.45 & 83.96 & 83.02 & 85.85 & 83.96 & 87.74 \\
monks\_3 & 90.41 & 90.41 & 85.63 & 91.02 & 74.85 & 85.03 & 95.81 \\
new-thyroid1.csv & 86 & 86 & 90.77 & 96.92 & 73.15 & 67.62 & 100 \\
oocytes\_merluccius\_nucleus\_4d & 80.71 & 79.74 & 80.78 & 78.5 & 67.43 & 77.2 & 84.69 \\
oocytes\_trisopterus\_nucleus\_2f & 80.12 & 80.94 & 75.55 & 80.29 & 69.71 & 71.9 & 83.21 \\
statlog\_australian\_credit & 68.75 & 68.75 & 62.02 & 62.02 & 70.19 & 70.19 & 69.71 \\
statlog\_german\_credit & 70.33 & 70.67 & 70 & 75.33 & 69 & 66.33 & 76.67 \\
vehicle2.csv & 90.85 & 90.03 & 91.73 & 92.52 & 90.36 & 85.9 & 98.43 \\
vertebral\_column\_2clases & 81.4 & 72.25 & 74.19 & 74.19 & 66.67 & 77.42 & 88.17 \\
votes.csv & 90.47 & 90.47 & 92.37 & 87.02 & 67.22 & 67.78 & 96.18 \\
vowel.csv & 99.33 & 98.99 & 85.19 & 63.97 & 70.71 & 67.03 & 99.66 \\
wpbc.csv & 69.49 & 70.97 & 76.27 & 69.49 & 75.69 & 84.96 & 81.36 \\
yeast-0-2-5-6\_vs\_3-7-8-9.csv & 89.71 & 89.71 & 93.38 & 90.73 & 77.16 & 67.02 & 93.05 \\
yeast-0-2-5-7-9\_vs\_3-6-8.csv & 95.35 & 95.68 & 98.01 & 97.35 & 78.44 & 70.71 & 98.34 \\
yeast-0-3-5-9\_vs\_7-8.csv & 88.82 & 90.79 & 69.74 & 45.39 & 70.01 & 81.01 & 89.47 \\
\hline 
Average Acc & \underline{81.55} & 80.81 & 79.62 & 78.17 & 75.86 & 76.62 & \textbf{85.2} \\ \hline
Average Rank & 3.88 & 4.34 & 4.07 & 4.38 & 4.95 & 4.36 & 2.03 \\
\hline
\multicolumn{8}{l}{The proposed model is denoted by $^{\dagger}$.}\\
 \multicolumn{8}{l}{The top and second-best models in terms of Acc are denoted by boldface and underline, respectively.}
\end{tabular}
}
\end{table*}

\section{Experimental Results}
To validate the performance of the proposed KRPRVFL model, extensive experiments are carried out on a collection of widely used benchmark datasets obtained from the UCI \cite{dua2017uci} and KEEL \cite{derrac2015keel} repositories. The proposed approach is thoroughly evaluated through comparative studies with several representative learning models, including RVFL \cite{pao1994learning}, ELM \cite{huang2006extreme}, GB-RVFL \cite{sajid2025gb}, GE-GB-RVFL \cite{sajid2025gb}, CRVFL \cite{liu2025complex}, and ACRVFL \cite{liu2025complex}, to assess its effectiveness comprehensively. Experiments with added label noise are detailed in Section S.I of the supplementary material. Sensitivity analyses of the proposed models are presented in Section S.II of the supplementary material.

\subsection{Experimental Setup}
Experiments are run on a Windows 11 workstation with an Intel Xeon Gold 6226R (2.90 GHz) and 256GB RAM using Python 3.11. Datasets are split 70:30 for training and testing, with hyperparameters tuned via grid search and five-fold cross-validation. The regularization coefficients are explored over the set $\mathcal{D} = \{10^{-5}, 10^{-4}, \ldots, 10^{5}\}$. The risk-sensitive parameter $\mu \in [1,10]$ and the power parameter $p \in [20,21,...,210]$. Hidden nodes  $N$ range from $3$ to $203$, and nine activation functions are tested: SELU, ReLU, Sigmoid, Sine, Hardlim, Tribas, Radbas, Sign, and Leaky ReLU.

\subsection{Evaluation on UCI and KEEL Datasets}
This section presents a comprehensive experimental study evaluating the proposed KRPRVFL model against multiple baseline models on 37 benchmark datasets from the UCI and KEEL repositories. Performance is assessed using classification accuracy (Acc), with results reported in Table \ref{Classification performance UCI and KEEL}. The proposed KRPRVFL model achieves an average Acc of $85.20\%$, while the baseline RVFL, ELM, GB-RVFL, GE-GB-RVFL, CRVFL, and ACRVFL models obtained an average Acc of $81.55\%$, $80.81\%$, $79.62\%$, $78.17\%$, $75.86\%$, and $76.62\%$, respectively. In terms of average Acc, the proposed KRPRVFL model consistently outperforms competing models, indicating strong predictive performance. However, average Acc alone may mask variability across datasets, as high performance on some can offset weaker results on others. To address this issue and to rigorously examine whether the observed performance variations are statistically meaningful, a series of nonparametric statistical tests is employed in accordance with the recommendations of \citet{demvsar2006statistical}. These statistical methods are suitable for comparing multiple classification algorithms across diverse datasets, particularly when parametric test assumptions are violated. Accordingly, nonparametric techniques such as ranking-based analysis, the Friedman test, and the Nemenyi post hoc test are used. In ranking-based evaluation, each model is ranked based on its performance on individual datasets, and the ranks are aggregated to determine overall performance. Under this ranking strategy, models with inferior performance are given larger rank scores, while more effective models obtain smaller ranks. This mechanism captures performance trade-offs across datasets, ensuring that strong results on some datasets can compensate for weaker outcomes on others. Consider the evaluation of $g$ models over $n$ datasets, where $\mathcal{R}_i^{j}$ denotes the rank assigned to the $j^{th}$ model on the $i^{th}$ dataset. The overall performance of the $j^{th}$ model is given by its mean rank, which is obtained by averaging its ranks across all datasets, i.e., $\mathcal{R}^{j} = \frac{1}{n} \sum_{i=1}^{n} \mathcal{R}_i^{j}.$ The proposed KRPRVFL model achieves an average rank of $2.03$, while the corresponding average ranks for the baseline RVFL, ELM, GB-RVFL, GE-GB-RVFL, CRVFL, and ACRVFL models are $3.88$, $4.34$, $4.07$, $4.38$, $4.95$, and $4.36$, respectively. Among all evaluated models, the proposed KRPRVFL attains the lowest average rank. Given that a lower rank indicates better performance, this result confirms that the proposed KRPRVFL model outperforms the baseline models. The Friedman test \cite{friedman1937use} is a nonparametric statistical test designed to examine whether meaningful performance differences exist among multiple models by evaluating their average rankings over a set of datasets. It offers a structured framework for comparing several models in multi-dataset experiments. The null hypothesis assumes that all models exhibit equivalent performance, implying that their average ranks are identical. The Friedman test statistic is calculated using a chi-square measure, denoted as $\chi_F^2$, which follows a chi-squared distribution with $g-1$ degrees of freedom, and is defined as: $\chi_F^2 = \frac{12n}{g(g+1)} \left[\sum_{j} \mathcal{R}_j^2 - \frac{g(g+1)^2}{4} \right].$ The Friedman statistic can be transformed into the $F_F$ statistic, given by $F_F = \frac{(n-1)\chi_F^2}{n(g-1) - \chi_F^2},$ which follows an $F$-distribution with $(g-1)$ and $(n-1)(g-1)$ degrees of freedom. For $n = 37$ datasets and $g = 7$ competing models, the Friedman analysis produces a test statistic of $\chi_F^2 = 41.807$, and the $F_F$ value is $8.35$. At the 5\% significance level, the corresponding critical value from the $F$-distribution for $F_F(6,216)$ is $2.1407$. As the obtained $F_F$ value is substantially larger than this threshold, the null hypothesis of equal performance is rejected, confirming the presence of statistically significant performance differences among the evaluated models. Now, the Nemenyi post hoc test is employed to analyze pairwise performance disparities between the competing models. The critical difference (C.D.) is determined as: $\text{C.D.} = q_{\alpha}\sqrt{\frac{g(g+1)}{6n}}$,
where $q_{\alpha}$ denotes the critical value obtained from the two-sided Nemenyi distribution table. Based on the $F$-distribution table, the critical value $q_\alpha$ at a 5\% significance level is $2.949$, which corresponds to a computed C.D. of $1.4811$. The average rank differences between KRPRVFL and RVFL, ELM, GB-RVFL, GE-GB-RVFL, CRVFL, and ACRVFL are $1.85$, $2.31$, $2.04$, $2.35$, $2.92$, and $2.33$, respectively. The Nemenyi test shows KRPRVFL achieves statistically significant improvements over all baselines.

\section{Conclusion}
In this paper, we introduced the KRPRVFL model, a robust extension of the random vector functional link network that integrates the kernel risk-sensitive mean p-power criterion with a collaborative learning mechanism. Our experiments on UCI and KEEL benchmark datasets demonstrated that KRPRVFL consistently achieves superior classification performance compared to existing RVFL variants and baseline models, particularly in the presence of noisy labels, outliers, and imbalanced data. The results highlight the effectiveness of risk-sensitive learning in improving robustness and the value of collaborative feature interaction for enhancing adaptability in complex data scenarios. KRPRVFL is computationally efficient and scalable, but its performance depends on kernel and risk-sensitive parameters and is best suited for small to medium datasets. Future work will target large-scale and streaming data, adaptive kernel selection, multi-task and multi-view learning, and deep RVFL integration for hierarchical features.

\bibliographystyle{IEEEtranN}
\bibliography{refs.bib}

\clearpage
\section*{Supplementary Material}

\renewcommand{\thesection}{S.I}
\section{Evaluation on UCI and KEEL datasets with added label noise}
The UCI and KEEL datasets reported in Table \ref{Classification performance Noise} represent realistic learning scenarios in which training data are often contaminated by label noise. Evaluating classification performance under such conditions is essential for verifying the robustness and reliability of learning algorithms. To this end, controlled label noise was injected at five different levels (5\%, 10\%, 20\%, 30\%, and 40\%) into several benchmark datasets, namely cleve, conn\_bench\_sonar\_mines\_rocks, ecoli-0-1-4-6\_vs\_5, haberman\_survival, and ionosphere. From the Table, it can be observed that the proposed KRPRVFL model consistently demonstrates strong robustness across different datasets and noise intensities. On the cleve dataset, KRPRVFL achieves the highest or near-highest classification Acc at most noise levels and attains the best average Acc among all competing models. For the conn\_bench\_sonar\_mines\_rocks dataset, KRPRVFL again shows competitive and stable behavior across all noise settings. While some baseline models experience notable performance fluctuations under higher noise levels, KRPRVFL maintains relatively consistent Acc and achieves the highest average Acc, demonstrating its ability to handle noisy decision boundaries effectively. On the ecoli-0-1-4-6\_vs\_5 dataset, which is particularly challenging due to class imbalance and sensitivity to noise, KRPRVFL significantly outperforms all baseline models. Even at higher noise levels, it preserves remarkably high classification Acc, resulting in a substantial margin in average Acc compared to other models. Similarly, on the ionosphere and haberman\_survival datasets, KRPRVFL consistently achieves superior performance across all noise ratios. While the Acc of baseline models declines sharply as noise increases, KRPRVFL retains a clear advantage, leading to the highest average Acc for this dataset. This further confirms its strong tolerance to noisy labels and its ability to preserve discriminative information. When considering the overall average Acc across all datasets and noise levels, KRPRVFL attains the best performance among all compared models. This overall superiority demonstrates that incorporating the KRP criterion into the RVFL framework substantially enhances robustness against label noise while maintaining strong generalization capability. 

\renewcommand{\thetable}{S.I}
\begin{table*}[ht!]
\centering
    \caption{Performance comparison of the proposed KRPRVFL model along with the baseline models for UCI and KEEL datasets with label noise.}
    \label{Classification performance Noise}
    \resizebox{1\linewidth}{!}{
\begin{tabular}{lcccccccc}
\hline
 Dataset & Noise & RVFL \cite{pao1994learning} & ELM \cite{huang2006extreme} & GB-RVFL \cite{sajid2025gb} & GE-GB-RVFL \cite{sajid2025gb} & CRVFL \cite{liu2025complex} & ACRVFL \cite{liu2025complex} & KRPRVFL$^{\dagger}$ \\
 \hline
cleve & 5\% & 75.11 & 70.78 & 70 & 73.33 & 70.56 & 70.56 & 80.78 \\
 & 10\% & 82.22 & 80 & 81.11 & 61.11 & 65.56 & 65.56 & 76.67 \\
 & 20\% & 80 & 75.11 & 75.56 & 66.67 & 65.56 & 64.89 & 81.11 \\
 & 30\% & 72.22 & 75.56 & 63.33 & 57.78 & 62.36 & 62.36 & 75.78 \\
 & 40\% & 57.78 & 58.89 & 66.67 & 52.22 & 45.56 & 45.56 & 58.89 \\ \hline
Average Acc &  & 73.47 & 72.07 & 71.33 & 62.22 & 61.92 & 61.79 & 74.64 \\ \hline
conn\_bench\_sonar\_mines\_rocks & 5\% & 76.19 & 63.49 & 71.43 & 76.19 & 66.67 & 68.25 & 80.95 \\
 & 10\% & 70.19 & 70.78 & 69.84 & 73.02 & 68.25 & 65.08 & 78.97 \\
 & 20\% & 70.95 & 76.19 & 74.6 & 77.78 & 69.84 & 69.84 & 71.43 \\
 & 30\% & 71.43 & 73.02 & 71.43 & 76.19 & 52.38 & 52.38 & 70.08 \\
 & 40\% & 66.67 & 52.38 & 69.84 & 69.84 & 41.27 & 42.86 & 69.97 \\ \hline
Average Acc &  & 71.09 & 67.17 & 71.43 & 74.6 & 59.68 & 59.68 & 74.28 \\ \hline
ecoli-0-1-4-6\_vs\_5 & 5\% & 90 & 90 & 83.33 & 94.05 & 90.95 & 90.95 & 98.81 \\
 & 10\% & 98.81 & 98.81 & 98.81 & 94.05 & 95.95 & 95.95 & 97.62 \\
 & 20\% & 98.81 & 98.81 & 98.81 & 100 & 95.95 & 95.95 & 96.43 \\
 & 30\% & 90.86 & 90.05 & 94.05 & 96.43 & 55.95 & 55.95 & 97.62 \\
 & 40\% & 64.05 & 61.67 & 65.48 & 58.33 & 45.95 & 45.95 & 94.05 \\ \hline
Average Acc &  & 88.51 & 87.87 & 88.1 & 88.57 & 76.95 & 76.95 & 96.9 \\ \hline
haberman\_survival & 5\% & 72.17 & 72.09 & 75 & 77.17 & 83.7 & 82.61 & 78.26 \\
 & 10\% & 77.17 & 75.35 & 76.09 & 77.17 & 79.35 & 73.91 & 75 \\
 & 20\% & 79.35 & 77.17 & 75.35 & 77.17 & 75 & 82.61 & 76.09 \\
 & 30\% & 51.09 & 59.35 & 76.09 & 61.96 & 82.61 & 80.43 & 72.52 \\
 & 40\% & 55.43 & 52.17 & 56.52 & 53.91 & 53.26 & 58.7 & 70.09 \\ \hline
Average Acc &  & 67.04 & 67.23 & 71.81 & 69.48 & 74.78 & 75.65 & 74.39 \\ \hline
ionosphere & 5\% & 85.68 & 85.74 & 86.79 & 83.02 & 83.02 & 85.85 & 86.79 \\
 & 10\% & 86.74 & 82.08 & 86.79 & 87.74 & 83.96 & 83.96 & 88.68 \\
 & 20\% & 85.58 & 83.96 & 82.74 & 81.13 & 83.02 & 82.08 & 87.74 \\
 & 30\% & 83.96 & 77.36 & 70.75 & 74.34 & 77.92 & 73.58 & 85.85 \\
 & 40\% & 55.6 & 53.77 & 58.49 & 55.66 & 47.55 & 40.81 & 65.09 \\ \hline
Average Acc &  & 79.51 & 76.58 & 77.11 & 76.38 & 75.09 & 73.26 & 82.83 \\ \hline
\multicolumn{2}{l}{Overall   Average Acc} & 75.92 & 74.18 & \underline{75.96} & 74.25 & 69.68 & 69.47 & \textbf{80.61} \\
\hline
\multicolumn{8}{l}{The proposed model is denoted by $^{\dagger}$.}\\
 \multicolumn{8}{l}{The top and second-best models in terms of Acc are denoted by boldface and underline, respectively.}
\end{tabular}
}
\end{table*}

\renewcommand{\thefigure}{S.1}
\begin{figure*}[ht!]
\begin{minipage}{.246\linewidth}
\centering
\subfloat[cleve]{\label{1a}\includegraphics[scale=0.24]{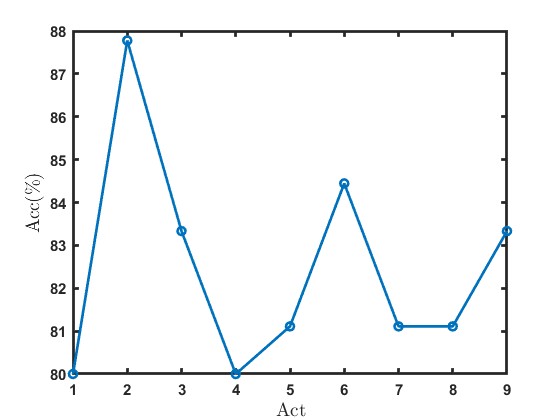}}
\end{minipage}
\begin{minipage}{.246\linewidth}
\centering
\subfloat[ecoli-0-1-4-6\_vs\_5]{\label{1b}\includegraphics[scale=0.24]{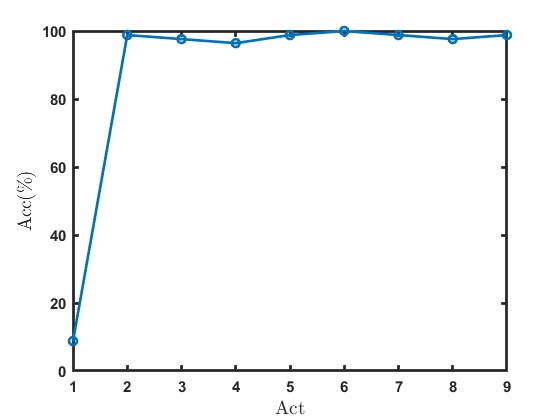}}
\end{minipage}
\begin{minipage}{.246\linewidth}
\centering
\subfloat[haberman\_survival]{\label{1c}\includegraphics[scale=0.24]{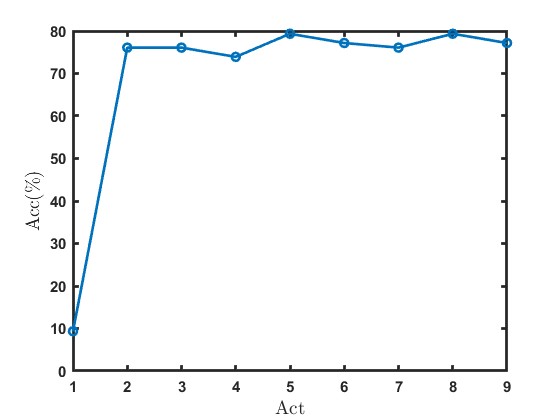}}
\end{minipage}
\begin{minipage}{.246\linewidth}
\centering
\subfloat[ionosphere]{\label{1d}\includegraphics[scale=0.24]{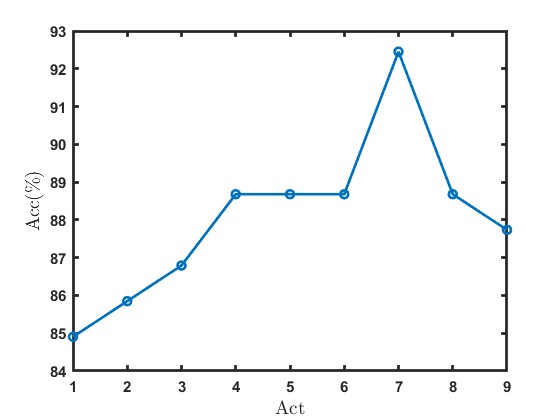}}
\end{minipage}
\caption{The impact of changing the activation function $\text(Act)$ on the Acc values of the proposed KRPRVFL model.}
\label{Effect of parameters Act}
\end{figure*}

\renewcommand{\thefigure}{S.2}
\begin{figure*}[ht!]
\begin{minipage}{.246\linewidth}
\centering
\subfloat[cleve]{\label{2a}\includegraphics[scale=0.24]{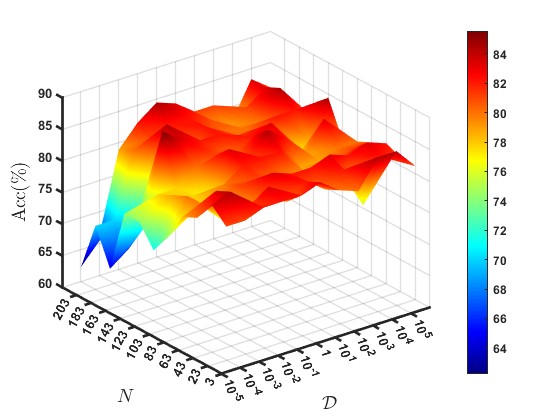}}
\end{minipage}
\begin{minipage}{.246\linewidth}
\centering
\subfloat[conn\_bench\_sonar\_mines\_rocks]{\label{2b}\includegraphics[scale=0.24]{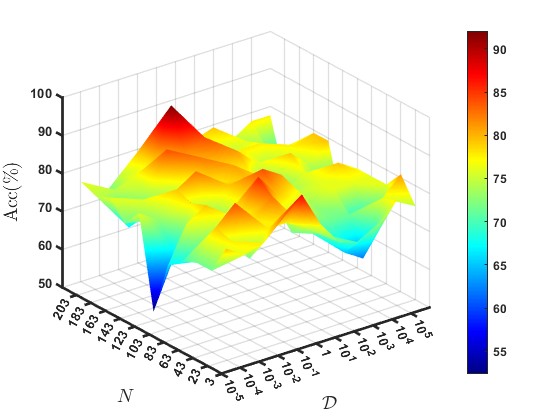}}
\end{minipage}
\begin{minipage}{.246\linewidth}
\centering
\subfloat[fertility]{\label{2c}\includegraphics[scale=0.24]{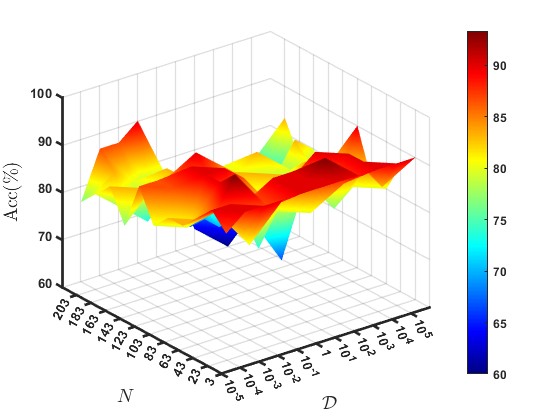}}
\end{minipage}
\begin{minipage}{.246\linewidth}
\centering
\subfloat[haberman\_survival]{\label{2d}\includegraphics[scale=0.24]{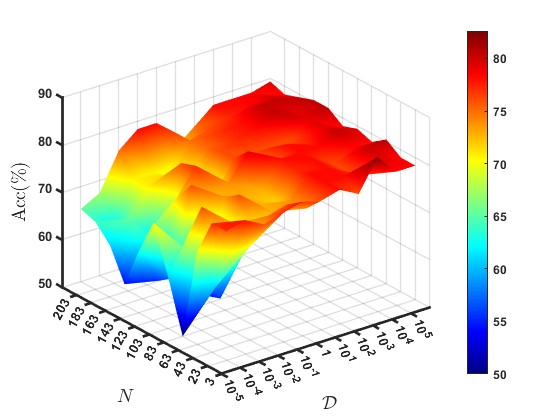}}
\end{minipage}
\caption{The impact of changing the parameters $\mathcal{D}$ and $N$ on the Acc values of the proposed KRPRVFL model.}
\label{Effect of parameters D and N}
\end{figure*}

\renewcommand{\thefigure}{S.3}
\begin{figure*}[ht!]
\begin{minipage}{.246\linewidth}
\centering
\subfloat[conn\_bench\_sonar\_mines\_rocks]{\label{3a}\includegraphics[scale=0.24]{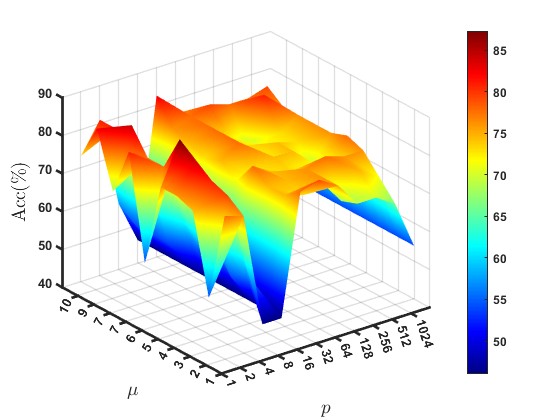}}
\end{minipage}
\begin{minipage}{.246\linewidth}
\centering
\subfloat[ecoli-0-1-4-6\_vs\_5]{\label{3b}\includegraphics[scale=0.24]{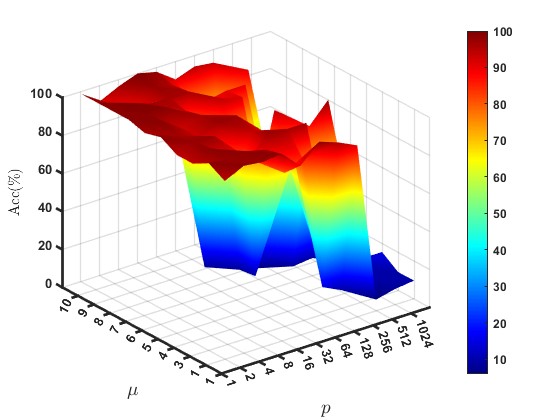}}
\end{minipage}
\begin{minipage}{.246\linewidth}
\centering
\subfloat[fertility]{\label{3c}\includegraphics[scale=0.24]{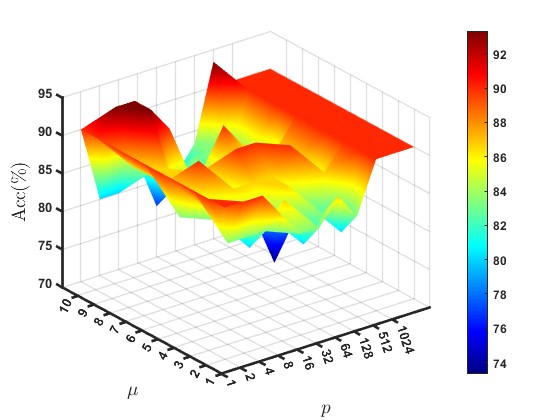}}
\end{minipage}
\begin{minipage}{.246\linewidth}
\centering
\subfloat[haberman\_survival]{\label{3d}\includegraphics[scale=0.24]{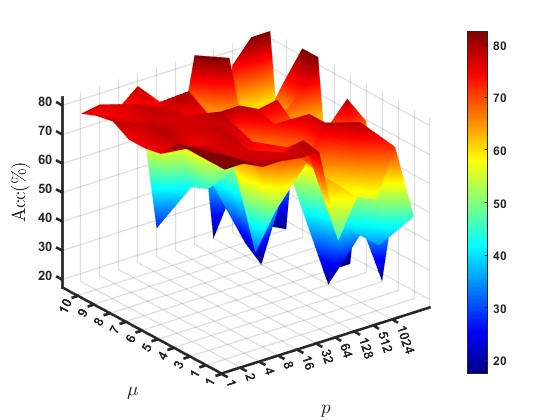}}
\end{minipage}
\caption{The impact of changing the parameters $p$ and $\mu$ on the Acc values of the proposed KRPRVFL model.}
\label{Effect of parameters p and mu}
\end{figure*}

\renewcommand{\thesection}{S.II}
\section{Sensitivity analysis}  
 This subsection examines the robustness and stability of the proposed KRPRVFL model with respect to its key hyperparameters. A comprehensive sensitivity analysis is carried out to investigate how changes in the activation function, network structure parameters, and robustness-related parameters affect the classification performance across different datasets. 
 
\subsubsection{Sensitivity analysis of the activation function (Act)}
Figure~\ref{Effect of parameters Act} illustrates the effect of different activation functions (Act) on the classification Acc of the proposed KRPRVFL model across four representative datasets. For the cleve dataset (Fig.~\ref{1a}), the Acc increases sharply when moving from SELU to ReLU, followed by moderate fluctuations for other activations, indicating a certain degree of sensitivity while maintaining overall stable performance. On the ecoli-0-1-4-6\_vs\_5 dataset (Fig.~\ref{1b}), the Acc remains consistently high across almost all activation functions, except for a noticeable drop with SELU, demonstrating strong robustness of the model to activation selection. In the case of the haberman\_survival dataset (Fig.~\ref{1c}), the performance shows mild variations, with Radbas and Tribas yielding slightly higher Acc, suggesting moderate dependence on the activation choice. Similarly, for the ionosphere dataset (Fig.~\ref{1d}), the model achieves its best performance with Radbas, while other activation functions also deliver competitive results. These observations indicate that KRPRVFL is largely robust to different activation functions, with radial-basis-type and ReLU-family activations providing consistently strong

\subsubsection{Sensitivity analysis of hyperparameters \texorpdfstring{$\mathcal{G}$}{G} and \texorpdfstring{$N$}{N}}
Figure~\ref{Effect of parameters D and N} illustrates the combined influence of the feature dimension parameter $\mathcal{D}$ and the number of enhancement nodes $N$ on the classification Acc of the proposed KRPRVFL model. For the cleve dataset (Fig.~\ref{2a}), higher Acc is generally achieved when both $\mathcal{D}$ and $N$ take moderate values, whereas very small configurations lead to noticeable performance degradation. In the conn\_bench\_sonar\_mines\_rocks dataset (Fig.~\ref{2b}), the Acc surface reveals increased sensitivity to $N$, with performance improving as $N$ increases, while extreme variations in $\mathcal{D}$ result in fluctuating behavior. For the fertility dataset (Fig.~\ref{2c}), optimal Acc is observed when $\mathcal{D}$ lies in a mid-range and $N$ is sufficiently large, indicating the importance of adequate model capacity for this dataset. Similarly, in the haberman\_survival dataset (Fig.~\ref{2d}), the model performs best under intermediate settings of both parameters, while overly small values of $\mathcal{D}$ or $N$ cause a clear drop in accuracy. Th results demonstrate that KRPRVFL maintains stable and competitive performance within a broad but well-defined region of $\mathcal{D}$ and $N$, highlighting its robustness and ease of parameter tuning in practical applications.

\subsubsection{Sensitivity analysis of hyperparameters \texorpdfstring{$p$}{p} and \texorpdfstring{$\mu$}{mu}}
Figure~\ref{Effect of parameters p and mu} illustrates the joint effect of the power parameter $p$ and the risk-sensitive parameter $\mu$ on the classification Acc of the proposed KRPRVFL model across four benchmark datasets. For the conn\_bench\_sonar\_mines\_rocks dataset (Fig.~\ref{3a}), the Acc surface shows noticeable fluctuations for small values of $p$, while more stable and higher performance is achieved when $p$ is set to moderate levels and $\mu$ takes intermediate values. In the ecoli-0-1-4-6\_vs\_5 dataset (Fig.~\ref{3b}), the model exhibits high Acc across a wide range of $\mu$ when $p$ is moderately large, whereas extreme combinations of small $p$ and large $\mu$ lead to sharp performance degradation. For the fertility dataset (Fig.~\ref{3c}), optimal results are obtained in a well-defined region where both $p$ and $\mu$ lie in their mid-ranges, indicating a balanced influence of robustness and sensitivity. Similarly, in the haberman\_survival dataset (Fig.~\ref{3d}), the proposed model achieves superior accuracy for intermediate parameter settings, while overly small or excessively large values of either $p$ or $\mu$ result in reduced performance. Hence, the observations demonstrate that KRPRVFL attains robust and stable classification performance within a broad yet structured region of the $(p,\mu)$ parameter space, underscoring the practical effectiveness and tunability of the proposed framework.

\end{document}